\PassOptionsToPackage{table}{xcolor}
\documentclass{article}
\usepackage[T1]{fontenc}
\usepackage{iclr2027_conference,times}
\usepackage{amsmath,amssymb}
\usepackage{graphicx}
\usepackage{needspace}
\usepackage{listings}
\usepackage{xcolor}
\usepackage{tikz}
\usepackage[most]{tcolorbox}
\usepackage{xcolor}
\usetikzlibrary{arrows.meta,positioning,fit,backgrounds}
\usepackage{hyperref}
\usepackage{listings}
\usepackage{xcolor}
\usepackage{enumitem}

\usepackage{float}
\usepackage{url}
\usepackage{graphicx}          
\usepackage[table]{xcolor}    
\usepackage{multirow}          
\usepackage{booktabs,graphicx}
\hypersetup{hidelinks}
\usepackage{booktabs}
\usepackage{graphicx}
\usepackage{array}
\usepackage{tabularx}

\hypersetup{
    colorlinks = true,
    linkcolor  = red,
    urlcolor   = magenta,
    citecolor  = cyan,
}

\newcommand{\method}{\textsc{SceneEvolve}}
\newcommand{\interface}{\textsc{SceneAct}}
\newcommand{\harness}{\textsc{SceneHarness-RSI}}

\title{Learning to Code the World: Harness Evolution for Compositional 3D Reconstruction}
\author{
    Haoyu Yao$^{1}$ \quad
    Chenguo Lin$^{1,\dagger}$ \quad
    Ruize Du$^{1}$ \quad
    Yadong Mu$^{1,\ddagger}$\\[0.5em]
    $^{1}$Peking University\\
    $^{\dagger}$Project lead; $^{\ddagger}$Corresponding author
}

\iclrfinalcopy
\begin{document}

\raggedbottom
\maketitle

\begin{figure}[h]
  \centering
  \includegraphics[width=0.95\linewidth]{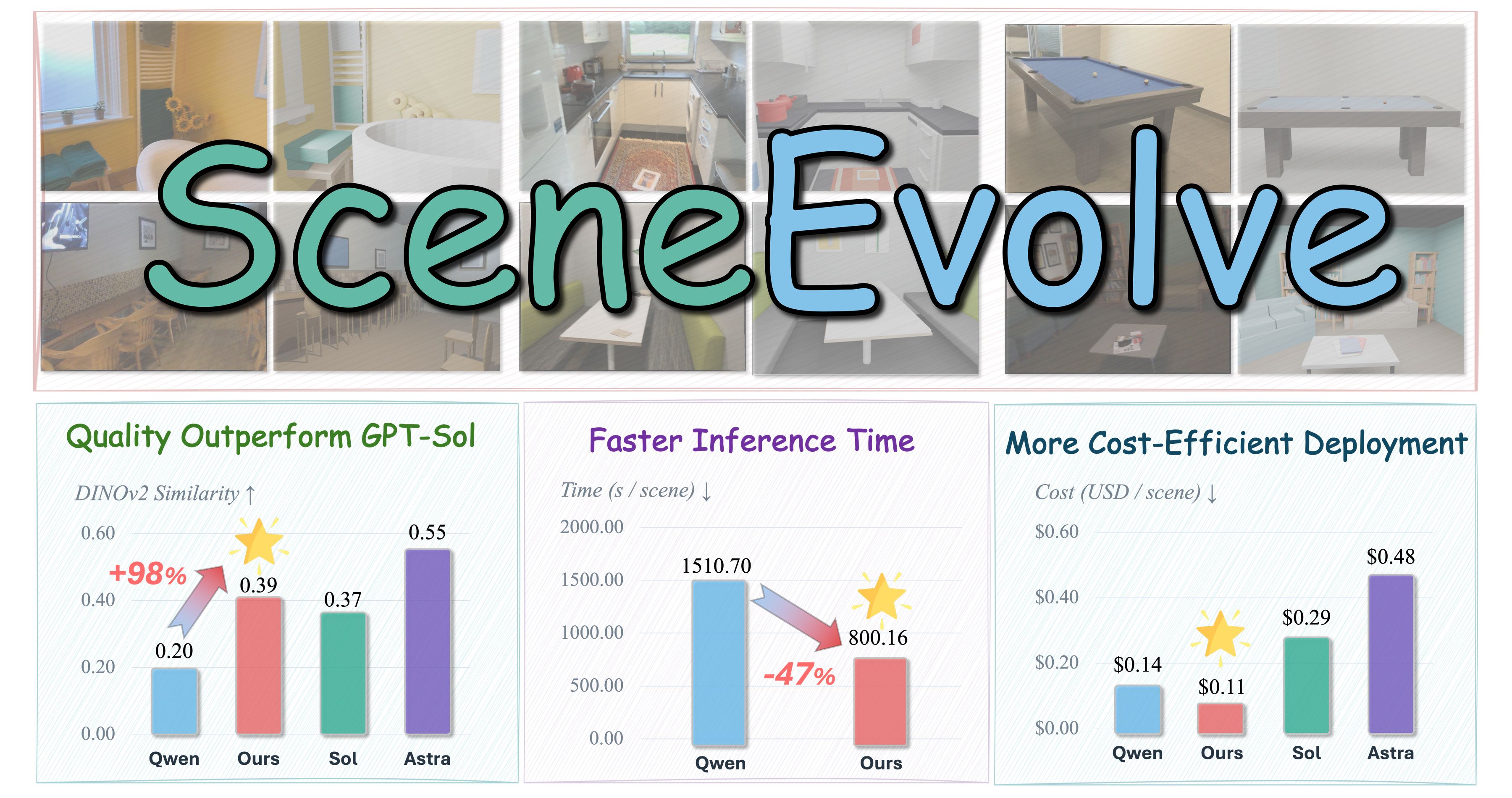}
  \caption{Our proposed SceneEvolve framework evolves the harness of a compact open-source code agent, enabling it to translate a single image into executable code that constructs an editable, compositional 3D scene. It improves reconstruction quality and reduces inference time compared with the base agent, while achieving higher reconstruction quality at lower cost than GPT-5.6-Sol.}
  \label{fig:opening}
\end{figure}

\begin{abstract}
    Reconstructing editable and compositional 3D scenes from a real image requires jointly reasoning about geometry, object structure, and spatial relationships. We introduce \method{}, a framework that represents 3D scenes as executable code and improves reconstruction by evolving the harness around a small open-source agent. \method{} combines \interface{}, a compositional semantic action interface built on parameterized primitives, hierarchical composition, and semantic spatial relations, with \harness{}, which leverages a stronger teacher agent to iteratively evolve the harness using feedback from development-set reconstructions. Throughout harness evolution, the model parameters, action semantics, and execution backend remain fixed. The resulting harness provides a reusable reconstruction procedure that enables the small open-source model to reconstruct unseen scenes without parameter updates or teacher assistance. Across two in-domain datasets and one out-of-domain dataset, \method{} with the comparatively compact Qwen3.8-27B model improves all four evaluated reconstruction metrics while reducing mean inference time by 47.0\% relative to the base model. More notably, despite relying on a substantially smaller model, \method{} achieves 32.3\% higher reconstruction quality than the large proprietary GPT-5.6-Sol model, while reducing mean equivalent deployment cost by 59.8\%. Furthermore, the harness evolved using Qwen3.8-27B also generalizes to Kimi-K3, achieving comparable reconstruction performance to GPT-6-Astra. The project page is \textbf{\url{https://haoyu-agent.github.io/SceneEvolve}}.

\end{abstract}

\section{Introduction}
\label{sec:introduction}
Compositional 3D scene reconstruction enables interactive and editable representations of real-world environments, with applications in robotics, embodied AI, AR/VR, and digital content creation~\citep{rest3d,scenecode,yeh2022photoscene}. Given a single image of an indoor environment, the goal is to reconstruct an editable, compositional 3D scene that captures its geometry, constituent objects, spatial layout, and lighting~\citep{rest3d,yeh2022photoscene}.

Existing approaches address this problem through several strategies. Procedural and generative methods synthesize diverse 3D environments~\citep{raistrick2024infinigenindoors,yang2024holodeck}, while multi-stage reconstruction pipelines divide the task into scene understanding, asset generation, and object placement~\citep{rest3d}. More recently, code-based agents have represented scenes as executable programs, enabling explicit manipulation of objects and spatial relations~\citep{viga,scenecode}. Despite these advances, reconstructing a specific real-world scene with high fidelity often requires carefully engineered intermediate stages~\citep{rest3d,yeh2022photoscene} or repeated render-and-refine iterations~\citep{viga,huang2024blenderalchemy}.

Frontier models such as GPT-6-Astra~\citep{gpt6astra} suggest a more direct approach: generating editable 3D scene programs from a single image with limited iterative refinement. However, their deployment costs motivate us to investigate whether a substantially smaller open-source model can achieve comparable image-to-3D programming capabilities through harness design rather than parameter updates~\citep{khattab2024dspy,agrawal2026gepa}. Our goal is to enable direct scene-code generation while reducing reliance on costly render-and-refine loops.

Achieving this goal presents two main challenges. First, compact models have limited visual and spatial reasoning capabilities, making it difficult to infer coherent 3D structures and translate visual observations into executable scene representations. Second, effective reconstruction also depends on the external harness: the instructions, tools, and control logic that govern how the model observes, plans, constructs, and executes a scene program~\citep{viga,khattab2024dspy}. Addressing these challenges requires both an interface that reduces the demands on geometric reasoning and a reusable procedure that guides reconstruction.

To address these challenges, we introduce \textbf{\method{}}, as shown in Figure~\ref{fig:opening}, a framework that combines a compositional scene interface, \interface{}, with a teacher-guided harness evolution framework, \harness{}. \interface{} translates high-level scene intent into executable operations through parameterized geometric primitives, hierarchical object composition, and semantic spatial relations. These abstractions reduce the need for low-level geometric programming while preserving explicit scene structure. Building on this interface, \harness{} uses a stronger teacher model to analyze the frozen student's development-set reconstructions, execution traces, and evaluation feedback, then propose revisions to the shared reconstruction harness. Candidate harnesses are evaluated under a common execution budget, and revisions are retained when they improve the development score. Throughout this process, the student parameters, action semantics, and execution backend remain fixed. The optimization target is the reusable reconstruction procedure, which is subsequently frozen and deployed on unseen scenes without teacher assistance.

Our contributions are summarized as follows:

\begin{itemize}[leftmargin=1cm,itemsep=0.2cm]
    \item We formulate compositional 3D scene reconstruction as an executable harness-optimization problem, targeting direct code generation without parameter updates or repeated iterations.

    \item We introduce \interface{}, a compositional interface that maps scene intent to executable 3D scene code using parameterized primitives, hierarchical composition, and semantic spatial relations, providing explicit control over object geometry and scene structure.

    \item We propose \harness{}, a teacher-guided recursive evolution method for optimizing reusable reconstruction harnesses. Teacher guidance is used only during evolution, allowing the resulting harnesses to generalize to unseen scenes without teacher assistance.

    \item Across in-domain and out-of-domain datasets, \method{} with Qwen3.8-27B outperforms \mbox{GPT-5.6-Sol}, reducing cost by 59.8\% and inference time by 47.0\% relative to the baseline. The evolved harness also transfers to Kimi-K3, achieving results competitive with \mbox{GPT-6-Astra} and demonstrating its transferability across different backbone models.
\end{itemize}

\section{Related Work}
\label{sec:related_work}

\paragraph{3D Generation for Compositional 3D Scene Reconstruction.}
Single-image reconstruction and generative models recover object geometry and appearance~\citep{hong2024lrm,xu2024instantmesh,xiang2025trellis}, while scene-level methods model object layouts and relations using semantic priors~\citep{instructscene,instructlayout} or jointly predict multiple objects and their spatial configurations~\citep{midi,scenegen}. \interface{} unifies object construction and scene composition in an executable program with primitives and spatial operators.

\paragraph{Multi-stage Pipelines for Compositional 3D Scene Reconstruction.}
Compositional reconstruction represents object geometry and scene layout as separately editable components. Prior work aligns CAD models or jointly estimates layout, poses, and geometry~\citep{izadinia2017im2cad,nie2020total3d}, with inverse rendering further recovering materials and lighting~\citep{yeh2022photoscene}. Recent methods combine single-image scene analysis, object generation, and spatial refinement~\citep{gen3dsr,rest3d,robosnap}, leverage multi-view evidence for assembly~\citep{simrecon,lucida}, or jointly generate object geometry and spatial configurations~\citep{midi,scenegen}. \method{} instead realizes object-centric reconstruction as an executable scene program built from parameterized primitives, part hierarchies, and spatial relations, enabling a frozen agent to construct and edit the scene.

\paragraph{Programmatic Coding Agents for Compositional 3D Scene Reconstruction.}
Procedural and language-guided generators use reusable assets and spatial constraints for controllable scene synthesis~\citep{deitke2022procthor,raistrick2024infinigenindoors,yang2024holodeck}. Related representations encode hierarchical assemblies as executable programs~\citep{shapeassembly} or predict structured scene descriptions from visual or point-cloud observations~\citep{avetisyan2024scenescript,spatiallm}. Visual coding agents further generate and refine graphics programs through rendered feedback, reusable libraries, and execution-guided iteration~\citep{hu2024scenecraft,huang2024blenderalchemy,viga,code_as_room,scenecode}. \interface{} exposes a fixed set of semantic construction operators, while \harness{} optimizes the reusable procedure.

\section{Methods}
\label{sec:methods}

\begin{figure}[t]
  \centering
  \includegraphics[width=\linewidth]{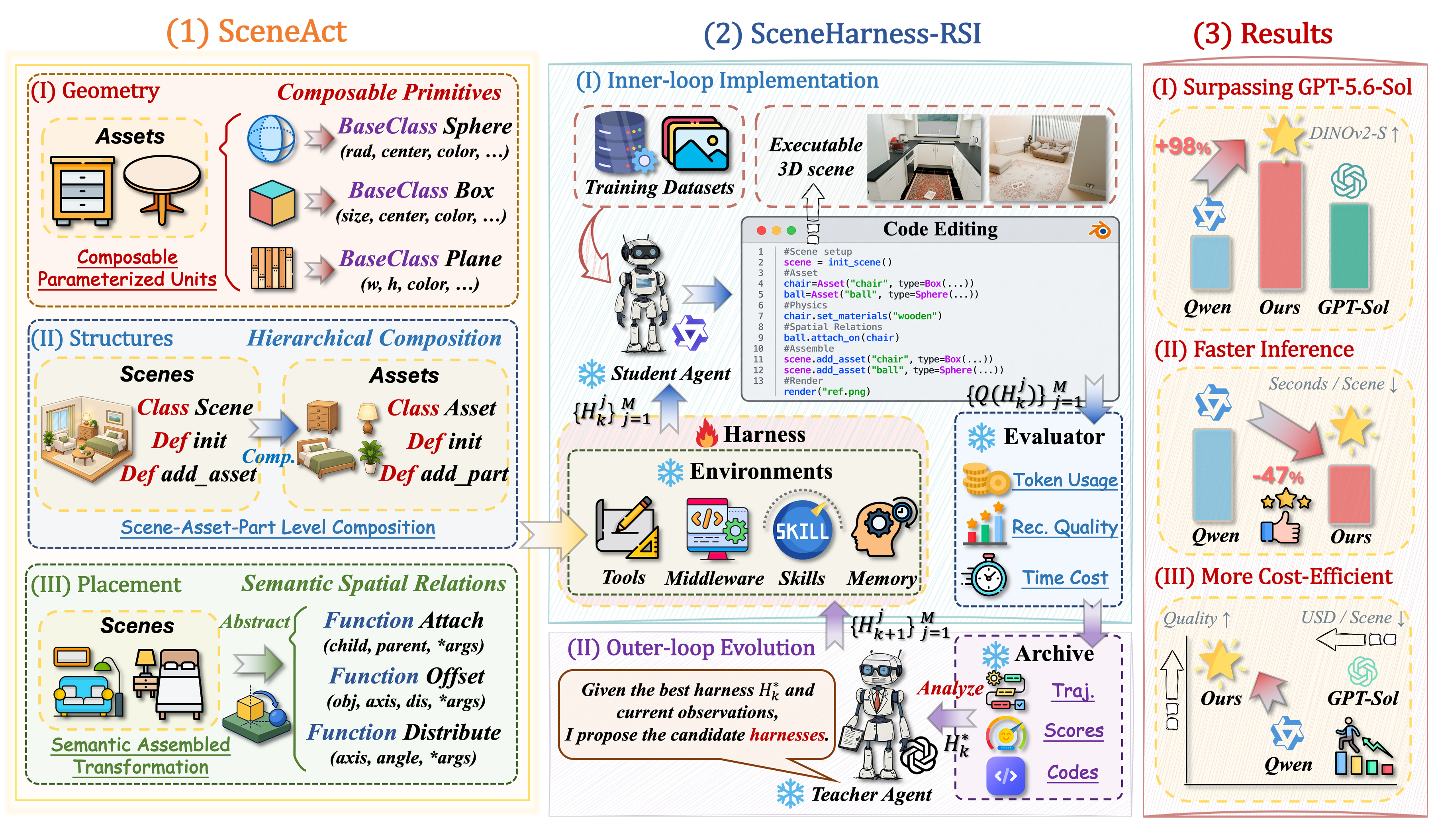}
  \vspace{-0.8cm}
  \caption{Illustration of \method{} for compositional image-to-3D reconstruction. The framework combines two components: (1) \interface{} translates scene intent into executable Blender operations through parameterized primitives, hierarchical composition, and semantic spatial relations. (2) \harness{} couples an outer loop for teacher-guided harness evolution with an inner loop for candidate evaluation by a frozen student, retaining revisions that improve performance. The optimized harness is deployed on unseen scenes without parameter updates.}
  \vspace{0.3cm}
  \label{fig:methods}
\end{figure}

\subsection{Overview}
\label{sec:overview}

We propose \method{}, a framework for compositional image-to-3D reconstruction of indoor scenes, as illustrated in Figure~\ref{fig:methods}. The framework comprises two components: \interface{}, a compact semantic interface that maps high-level scene intent to executable actions, and \harness{}, a teacher-guided procedure that ens the reconstruction harness using end-to-end reconstruction feedback. \interface{} defines the executable interface through which the frozen student agent reconstructs 3D scenes. \harness{} iteratively refines the harness through a propose--execute--evaluate--retain cycle to improve how the student uses \interface{} while keeping the student parameters fixed.

\subsection{SceneAct: A Compact Semantic Interface from Intent to Actions}
\label{sec:sceneact}

\interface{} provides a compact semantic interface between high-level scene reasoning and low-level Blender execution. It enables the student agent to use semantically meaningful actions that are deterministically translated into executable Blender operations, reducing the need to directly specify low-level geometric operations and numerical parameters. The action space comprises three complementary abstractions: \emph{composable parameterized primitives}, \emph{hierarchical composition}, and \emph{semantic spatial relations}. Together, these abstractions support progressive scene construction from local geometry to object structure and scene-level layout.

\paragraph{Composable Parameterized Primitives.}
\interface{} exposes a compact set of parameterized geometric primitives as the basic units of scene construction. A \interface{} program can instantiate multiple primitives, with each primitive $\rho_i$ represented as
\begin{equation}
\rho_i
=
\bigl(
\kappa_i,
\mathbf{g}_i,
\mathbf{a}_i
\bigr),
\qquad
\kappa_i \in \mathcal{G},
\label{eq:sceneact_primitive}
\end{equation}
where $\kappa_i$ identifies a primitive class in the predefined library $\mathcal{G}$, and $\mathbf{g}_i$ and $\mathbf{a}_i$ denote its geometric and appearance parameters, respectively. This representation allows the student agent to construct geometry by instantiating and configuring primitive classes without directly manipulating mesh vertices, faces, or other low-level Blender representations.

The primitive classes share a common interface for translation, rotation, and scaling, allowing their instances to be composed into higher-level scene structures. For example, a tabletop can be represented by a thin \emph{Box}, its supporting legs by \emph{Box} or \emph{Cylinder} instances, and a curved lamp arm by a \emph{Curve}. These primitives can then be transformed and assembled into semantic parts and assets. This design expresses reconstruction as a sequence of compact, executable decisions and shifts the student's reasoning from low-level mesh programming to structured geometric construction. In this way, the student's geometric reasoning is reduced to two interpretable decisions: \emph{which primitive best explains each visible part, and what geometric and appearance parameters characterize it?}

\paragraph{Hierarchical Composition.}
We organize the reconstructed scene using an explicit semantic hierarchy implemented in the scene program. The \emph{Scene} class manages a collection of assets through interfaces such as \emph{scene.add\_asset(...)}. The reusable \emph{Asset} base class encapsulates object-level structure and provides \emph{asset.add\_part(...)} for composing semantic parts. Each \emph{Part} is constructed from one or more parameterized geometric primitives. This hierarchy separates scene-level composition, object-level structure, and part-level geometry: the \emph{Scene} supports adding, removing, transforming, and duplicating complete assets, while each \emph{Asset} controls the construction and refinement of its internal parts. For example, a table asset may contain a tabletop and several legs, each constructed independently from geometric primitives. This program structure supports modular construction and localized editing while preserving a direct correspondence between visually identifiable entities in the input image and their executable representations.

\paragraph{Semantic Spatial Relations.}
\interface{} specifies spatial structure through semantic relations, reducing the need to predict explicit coordinates. This design is motivated by the observation that VLMs are more reliable at inferring relations such as \emph{under}, \emph{attached to}, and \emph{facing} than at predicting the precise 3D translations and rotations needed to realize them. Accordingly, each scene element exposes geometric features with semantic meaning, including \emph{center}, \emph{extent}, \emph{support}, \emph{anchor}, and \emph{boundary}, together with relational operators such as \emph{align(...)} and \emph{distribute(...)}.

These operators encode spatial constraints relative to scene geometry. For example, four table legs can be positioned below a tabletop and aligned with its corners without predicting their coordinates independently. The \interface{} interpreter resolves such relations into deterministic transformations based on current geometry, reference frames, extents, and anchors. Since the constraints are defined structurally rather than in fixed world coordinates, they can be re-evaluated after edits, allowing dependent elements to update automatically. This separates relational reasoning from numerical transformation computation, reducing direct coordinate prediction while preserving precision.

\subsection{\harness{}: A teacher-guided recursive harness evolution}
\label{sec:harness_evolution}

\harness{} improves the execution harness of a frozen student through teacher-guided revisions. Starting from an initial harness $H_0$, the student reconstructs development-set scenes through \interface{}, producing Blender programs, rendered outputs, and execution traces. A teacher uses this evidence to identify limitations of the current harness and propose targeted revisions. The same frozen student then evaluates the revised harnesses under a fixed protocol, and only revisions that improve the recorded development-set reconstruction score are retained. Repeating this propose--execute--evaluate--retain cycle defines an outer-loop search over executable harnesses.

\paragraph{Harness Search Space and Evaluation.}
The harness defines the execution environment of the frozen student, including callable tool interfaces, middleware, reusable skills, and memory. \harness{} searches over revisions to these components while keeping the \interface{} semantics, student parameters, and Blender execution backend fixed.

Let $\mathcal{D}_{\mathrm{dev}}=\{I_i\}_{i=1}^{N}$ denote a development set of $N$ images. For each image $I_i$, the student executes under harness $H$ to produce a reconstructed scene and its rendered image $\hat{I}_i(H)$. We define the development-set quality score $Q(H)$ as the average of three measures across development images:
\begin{equation}
Q(H)=\frac{1}{3N}
\sum_{i=1}^{N}
\left[
\mathcal{S}\left(I_i,\hat{I}_i(H)\right)
+
\left(1-\mathcal{L}\left(I_i,\hat{I}_i(H)\right)\right)
+
\frac{1+\mathcal{D}\left(I_i,\hat{I}_i(H)\right)}{2}
\right],
\label{eq}
\end{equation}
where $\mathcal{S}$, $\mathcal{L}$, and $\mathcal{D}$ denote SSIM~\citep{wang2004ssim}, LPIPS~\citep{zhang2018lpips}, and DINOv2 cosine similarity~\citep{oquab2023dinov2}, respectively. Before aggregation, we clip $\mathcal{S}$ and $\mathcal{L}$ to $[0,1]$ and $\mathcal{D}$ to $[-1,1]$. The resulting transformations place all three terms on a common $[0,1]$ scale. Thus, a higher score $Q(H)\in[0,1]$ indicates better reconstruction quality. We use this score to compare and select harness candidates.

\paragraph{Outer-Loop Teacher-Guided Harness Evolution.}
We search the discrete space of executable harnesses using teacher-guided proposals. At round $k$, the search maintains the retained harness $H_k$ and an archive $\mathcal A_k$ containing all previously evaluated harnesses, their recorded development scores, and their execution traces. The teacher examines this history to identify recurring reconstruction and execution failures and proposes $M$ candidate revisions:
\begin{equation}
H_{k+1}^{j}
\sim
P_{\phi}\!\left(\,\cdot \mid H_k,\mathcal A_k\right),
\qquad j=1,\ldots,M.
\label{eq:harness_proposal}
\end{equation}
Here, $P_{\phi}$ denotes the proposal distribution induced by the teacher model with parameters $\phi$. Revisions may include improved diagnostics, changes to tool handling, and reusable routines that apply \interface{} operators. These revisions change how the student performs reconstruction while preserving the operator semantics and student parameters.

\paragraph{Inner-Loop Student Evaluation and Harness Retention.}
The same frozen student evaluates each candidate harness on $\mathcal D_{\mathrm{dev}}$ under identical decoding configurations, rollout budgets, rendering settings, and scoring protocols. Let $H_{k+1}^{*}$ denote a highest-scoring candidate in the current round. We replace the retained harness only when this candidate achieves a score strictly higher than the retained harness's recorded development score:
\begin{equation}
H_{k+1}
=
\begin{cases}
H_{k+1}^{*},
&
Q\!\left(H_{k+1}^{*}\right)
>
Q(H_k), \\[2pt]
H_k,
&
\text{otherwise}.
\end{cases}
\label{eq:harness_retention}
\end{equation}
All evaluated candidates, their scores, and their execution traces are added to the archive, allowing subsequent proposals to draw on both successful and unsuccessful revisions. This retention rule ensures that the recorded development scores of the retained harnesses form a non-decreasing sequence. It does not guarantee monotonic improvement in expected performance or performance on unseen scenes. Student reconstruction forms the inner execution loop, while teacher-guided harness revision forms the outer search loop. After $K$ rounds, the method returns the retained harness $H_K$.

\section{Experiments}

\begin{table}[t]
\centering
\caption{Performance comparison across different models on ScanNet and CA-1M (in-domain) and ARKitScenes (out-of-domain). We report PSNR, SSIM, LPIPS, and DINOv2 similarity. GPT-6-Astra serves as the reference. The best and second results are highlighted in \textbf{bold} and \underline{underlined}.}
\label{tab:bvb-image-metrics}
\begingroup
\setlength{\tabcolsep}{2.2pt}
\renewcommand{\arraystretch}{1.4}
\setlength{\aboverulesep}{0.45ex}
\setlength{\belowrulesep}{0.55ex}
\setlength{\heavyrulewidth}{1.2pt}
\setlength{\lightrulewidth}{0.6pt}
\setlength{\cmidrulewidth}{0.6pt}
\IfFileExists{./figures/gpt-teal.png}{
  \def\ModelLogoPath{./figures/}
}{\def\ModelLogoPath{figures/}}
\newcommand{\ModelIcon}[1]{
  \makebox[1.55em][c]{\raisebox{-0.22em}{
    \includegraphics[width=1.25em,height=1.25em,keepaspectratio]{\ModelLogoPath#1}}}
}
\newcommand{\GPTIcon}{\ModelIcon{gpt-teal.png}}
\newcommand{\ClaudeIcon}{\ModelIcon{claude-orange.png}}
\newcommand{\GeminiIcon}{\ModelIcon{gemini-blue.png}}
\newcommand{\QwenIcon}
{\ModelIcon{qwen-violet.png}}
\newcommand{\KimiIcon}
{\ModelIcon{kimi.png}}
\newcommand{\TableGain}[1]{\textcolor{green!35!black}{\textbf{#1}}}
\newcommand{\TableLoss}[1]{\textcolor{red!65!black}{#1}}
\resizebox{\linewidth}{!}{
\begin{tabular}{lc*{12}{c}}
\toprule[1.4pt]
\multicolumn{2}{c}{}
& \multicolumn{8}{c}{\textbf{In-domain}}
& \multicolumn{4}{c}{\textbf{Out-of-domain}} \\
\cmidrule[1.0pt](lr){3-10}
\cmidrule[1.0pt](lr){11-14}
\multicolumn{1}{c}{\textbf{Model}}
& \shortstack{\textbf{Inference effort}}
& \multicolumn{4}{c}{\textbf{ScanNet}}
& \multicolumn{4}{c}{\textbf{CA-1M}}
& \multicolumn{4}{c}{\textbf{ARKitScenes}} \\
\cmidrule[0.8pt](lr){3-6}
\cmidrule[0.8pt](lr){7-10}
\cmidrule[0.8pt](lr){11-14}
&
& PSNR$\uparrow$ & SSIM$\uparrow$ & LPIPS$\downarrow$ & DINOv2-S$\uparrow$
& PSNR$\uparrow$ & SSIM$\uparrow$ & LPIPS$\downarrow$ & DINOv2-S$\uparrow$
& PSNR$\uparrow$ & SSIM$\uparrow$ & LPIPS$\downarrow$ & DINOv2-S$\uparrow$ \\
\midrule[1.2pt]
\rowcolor{teal!6}
\GPTIcon\ {GPT-6-Astra}
& High
& \textcolor{gray}{10.063} & \textcolor{gray}{0.386} & \textcolor{gray}{0.684} & \textcolor{gray}{0.590} & \textcolor{gray}{10.309} & \textcolor{gray}{0.382} & \textcolor{gray}{0.705} & \textcolor{gray}{0.556} & \textcolor{gray}{9.561} & \textcolor{gray}{0.407} & \textcolor{gray}{0.748} & \textcolor{gray}{0.519} \\
\rowcolor{teal!3}
\GPTIcon\ {GPT-6-Astra}
& Medium
& \textcolor{gray}{10.143} & \textcolor{gray}{0.383} & \textcolor{gray}{0.684} & \textcolor{gray}{0.579} & \textcolor{gray}{10.433} & \textcolor{gray}{0.378} & \textcolor{gray}{0.711} & \textcolor{gray}{0.557} & \textcolor{gray}{10.177} & \textcolor{gray}{0.423} & \textcolor{gray}{0.732} & \textcolor{gray}{0.528} \\
\GPTIcon\ {GPT-6-Astra}
& Low
& \textcolor{gray}{9.888} & \textcolor{gray}{0.385} & \textcolor{gray}{0.701} & \textcolor{gray}{0.596} & \textcolor{gray}{10.108} & \textcolor{gray}{0.372} & \textcolor{gray}{0.691} & \textcolor{gray}{0.552} & \textcolor{gray}{10.245} & \textcolor{gray}{0.403} & \textcolor{gray}{0.712} & \textcolor{gray}{0.475} \\
\addlinespace[2pt]\midrule[1.0pt]\addlinespace[1pt]
\rowcolor{orange!6}
\ClaudeIcon\ {Claude-Opus-5}
& High
& 9.174 & 0.352 & 0.735 & 0.438 & 8.781 & 0.368 & 0.749 & 0.441 & 8.551 & 0.389 & 0.751 & \underline{0.417} \\
\rowcolor{orange!3}
\ClaudeIcon\ {Claude-Opus-5}
& Medium
& 9.021 & 0.348 & 0.741 & \underline{0.444} & 8.705 & 0.361 & 0.756 & 0.432 & 8.498 & 0.392 & 0.754 & 0.409 \\
\ClaudeIcon\ {Claude-Opus-5}
& Low
& 8.742 & 0.337 & 0.747 & 0.416 & 8.566 & 0.355 & 0.753 & 0.419 & 8.287 & 0.374 & 0.764 & 0.391 \\
\ClaudeIcon\ {Claude-Opus-5}
& None
& 8.463 & 0.329 & 0.759 & 0.398 & 8.318 & 0.346 & 0.769 & 0.423 & 8.146 & 0.358 & 0.771 & 0.356 \\
\addlinespace[2pt]\midrule[1.0pt]\addlinespace[1pt]
\rowcolor{teal!6}
\GPTIcon\ {GPT-5.6-Sol}
& High
& 9.042 & 0.337 & 0.742 & 0.417 & 8.562 & 0.353 & 0.748 & 0.428 & 8.312 & 0.371 & 0.755 & 0.401 \\
\rowcolor{teal!3}
\GPTIcon\ {GPT-5.6-Sol}
& Medium
& 8.731 & 0.319 & 0.754 & 0.375 & 7.821 & 0.347 & 0.765 & 0.411 & 8.027 & 0.340 & 0.752 & 0.408 \\
\GPTIcon\ {GPT-5.6-Sol}
& Low
& 7.945 & 0.307 & 0.756 & 0.382 & 8.294 & 0.338 & 0.757 & 0.416 & 7.942 & 0.314 & 0.776 & 0.349 \\
\GPTIcon\ {GPT-5.6-Sol}
& None
& 7.755 & 0.284 & 0.752 & 0.389 & 8.212 & 0.321 & 0.770 & 0.361 & 8.067 & 0.356 & 0.773 & 0.357 \\
\addlinespace[2pt]\midrule[1.0pt]\addlinespace[1pt]
\rowcolor{blue!6}
\GeminiIcon\ {Gemini-Flash-3.8}
& High
& 8.126 & 0.318 & 0.774 & 0.358 & 8.113 & 0.324 & 0.781 & 0.365 & 7.984 & 0.337 & 0.785 & 0.341 \\
\rowcolor{blue!3}
\GeminiIcon\ {Gemini-Flash-3.8}
& Medium
& 8.004 & 0.311 & 0.782 & 0.361 & 8.154 & 0.319 & 0.779 & 0.358 & 7.846 & 0.330 & 0.792 & 0.334 \\
\GeminiIcon\ {Gemini-Flash-3.8}
& Low
& 7.813 & 0.298 & 0.789 & 0.337 & 7.874 & 0.306 & 0.795 & 0.341 & 7.731 & 0.332 & 0.799 & 0.336 \\
\addlinespace[2pt]\midrule[1.0pt]\addlinespace[1pt]
\rowcolor{violet!6}
\QwenIcon\ {Qwen3.8-27B}
& xHigh
& 9.061 & 0.376 & 0.772 & 0.240 & 8.381 & 0.367 & 0.806 & 0.134 & 9.704 & 0.448 & 0.785 & 0.219 \\
\addlinespace[2pt]
\rowcolor{violet!16}
\QwenIcon\ \textbf{Qwen3.8-27B w/ Ours}
& xHigh
& 9.124 & \underline{0.378} & 0.755 & 0.391 & 8.513 & \underline{0.440} & \underline{0.742} & 0.418 & \underline{9.862} & \underline{0.453} & \underline{0.721} & 0.368 \\
\rowcolor{violet!3}
\textbf{\textit{Improvement}}
& - 
& \textcolor{green!35!black}{+0.70\%} & \textcolor{green!35!black}{+0.53\%} & \textcolor{green!35!black}{-2.20\%} & \textcolor{green!35!black}{+62.92\%} & \textcolor{green!35!black}{+1.57\%} & \textcolor{green!35!black}{+19.89\%} & \textcolor{green!35!black}{-7.94\%} & \textcolor{green!35!black}{+211.94\%} & \textcolor{green!35!black}{+1.63\%} & \textcolor{green!35!black}{+1.12\%} & \textcolor{green!35!black}{-8.15\%} & \textcolor{green!35!black}{+68.04\%} \\
\addlinespace[2pt]\midrule[1.0pt]\addlinespace[1pt]
\rowcolor{red!8}
\KimiIcon\ Kimi-K3
& High
& \underline{9.181} & 0.350 & \underline{0.734} & 0.437 & \underline{8.847} & 0.365 & 0.756 & \underline{0.458} & 8.603 & 0.384 & 0.758 & 0.392 \\
\rowcolor{red!16}
\KimiIcon\ \textbf{Kimi-K3 w/ Ours}
& High
& \textbf{9.547} & \textbf{0.381} & \textbf{0.712} & \textbf{0.513} & \textbf{9.718} & \textbf{0.454} & \textbf{0.722} & \textbf{0.534} & \textbf{10.029} & \textbf{0.469} & \textbf{0.716} & \textbf{0.504} \\
\rowcolor{red!4}
\textbf{\textit{Improvement}}
& - 
& \textcolor{green!35!black}{+3.99\%} & \textcolor{green!35!black}{+8.86\%} & \textcolor{green!35!black}{-3.00\%} & \textcolor{green!35!black}{+17.39\%} & \textcolor{green!35!black}{+9.85\%} & \textcolor{green!35!black}{+24.38\%} & \textcolor{green!35!black}{-4.50\%} & \textcolor{green!35!black}{+16.59\%} & \textcolor{green!35!black}{+16.58\%} & \textcolor{green!35!black}{+22.14\%} & \textcolor{green!35!black}{-5.54\%} & \textcolor{green!35!black}{+28.57\%} \\

\bottomrule[1.4pt]
\end{tabular}
}
\endgroup
\vspace{0.2cm}
\end{table}

\subsection{Experimental Settings}

\paragraph{Datasets.}
We use three indoor scene datasets comprising 160 scenes in total: ScanNet~\citep{dai2017scannet} and CA-1M~\citep{lazarow2024cubify} for in-domain development and evaluation, and ARKitScenes~\citep{baruch2021arkitscenes} for out-of-domain evaluation. The resulting split consists of 100 development scenes, 40 in-domain test scenes, and 20 out-of-domain test scenes.

\vspace{-0.2cm}
\paragraph{In-domain Datasets.} ScanNet~\citep{dai2017scannet} provides RGB-D sequences with camera poses, reconstructed indoor geometry, and semantic annotations, while CA-1M~\citep{lazarow2024cubify} contains handheld egocentric captures registered to high-quality laser scans with dense 3D object annotations. From each dataset, we randomly sample 70 scenes and partition them at the scene level into disjoint development and test sets of 50 and 20 scenes, respectively. All views and repeated captures of the same physical scene are confined to a single split. The development sets are used for recursive harness search, while the test sets are reserved exclusively for in-domain evaluation.

\vspace{-0.2cm}
\paragraph{Out-of-domain Dataset.}
ARKitScenes~\citep{baruch2021arkitscenes} contains mobile RGB-D indoor captures with laser-scanned geometry and 3D object annotations. We randomly sample 20 scenes for out-of-domain evaluation. The harness developed on ScanNet~\citep{dai2017scannet} and CA-1M~\citep{lazarow2024cubify} is transferred directly to these scenes with the student weights, harness code, and inference configuration fixed. No ARKitScenes~\citep{baruch2021arkitscenes} data or feedback are used during harness development, enabling a clean evaluation of cross-dataset generalization.

\vspace{-0.2cm}
\paragraph{Evaluation Metrics.} We evaluate all methods along three dimensions: reconstruction quality, inference efficiency, and equivalent cost. Reconstruction quality is measured by PSNR~\citep{wang2009mse}, SSIM~\citep{wang2004ssim}, LPIPS~\citep{zhang2018lpips}, and DINOv2 similarity~\citep{oquab2023dinov2}, capturing pixel, structural, perceptual, and semantic fidelity, respectively. Inference efficiency is measured by the average per-scene inference time under a unified protocol. We further report the average equivalent cost per scene under a consistent cost definition, enabling direct comparison of the trade-offs among quality, efficiency, and cost across model configurations.

\vspace{-0.2cm}
\paragraph{Baselines.} We benchmark proprietary multimodal models as zero-shot agents for compositional image-to-3D scene reconstruction, without task-specific fine-tuning. Our baseline suite includes GPT-5.6-Sol~\citep{gpt56sol} and GPT-6-Astra~\citep{gpt6astra}, Claude-Opus-5~\citep{claudeopus5}, and Gemini-Flash-3.8~\citep{geminiflash38}. For models with configurable reasoning effort, we evaluate all supported settings among \emph{none}, \emph{low}, \emph{medium}, and \emph{high}, treating each model--effort pair as a distinct baseline configuration. We also evaluate Qwen3.8-27B~\citep{qwen38_27b} at \emph{xhigh}, using the same model and reasoning configuration as the frozen student in our method. We additionally use Kimi-K3~\citep{kimik3} at \emph{high} as a cross-model validation harness.

\vspace{-0.2cm}
\paragraph{Implementation details.} We use Qwen3.8-27B~\citep{qwen38_27b} at \emph{xhigh} as the frozen student and GPT-6-Astra~\citep{gpt6astra} at \emph{medium} as the harness proposer, running $K=30$ rounds with $M=2$ proposals each. Candidates are conditioned on archived development evidence and evaluated on equally weighted ScanNet~\citep{dai2017scannet} and CA-1M~\citep{lazarow2024cubify} scenes, with input frames selected by a space-aware frame sampling strategy following Spatial-MLLM~\citep{wu2025spatialmllm}. The selected harness and inference configuration are frozen at test. All models are accessed exclusively via API calls, without using additional third-party harnesses such as Codex, Claude Code, or pi. To evaluate the cross-model transferability of the self-evolved harness, we apply the harness evolved with Qwen3.8-27B directly to Kimi-K3~\citep{kimik3} at \emph{high}.

\begin{figure}[!t]
  \centering
  \includegraphics[width=0.97\linewidth]{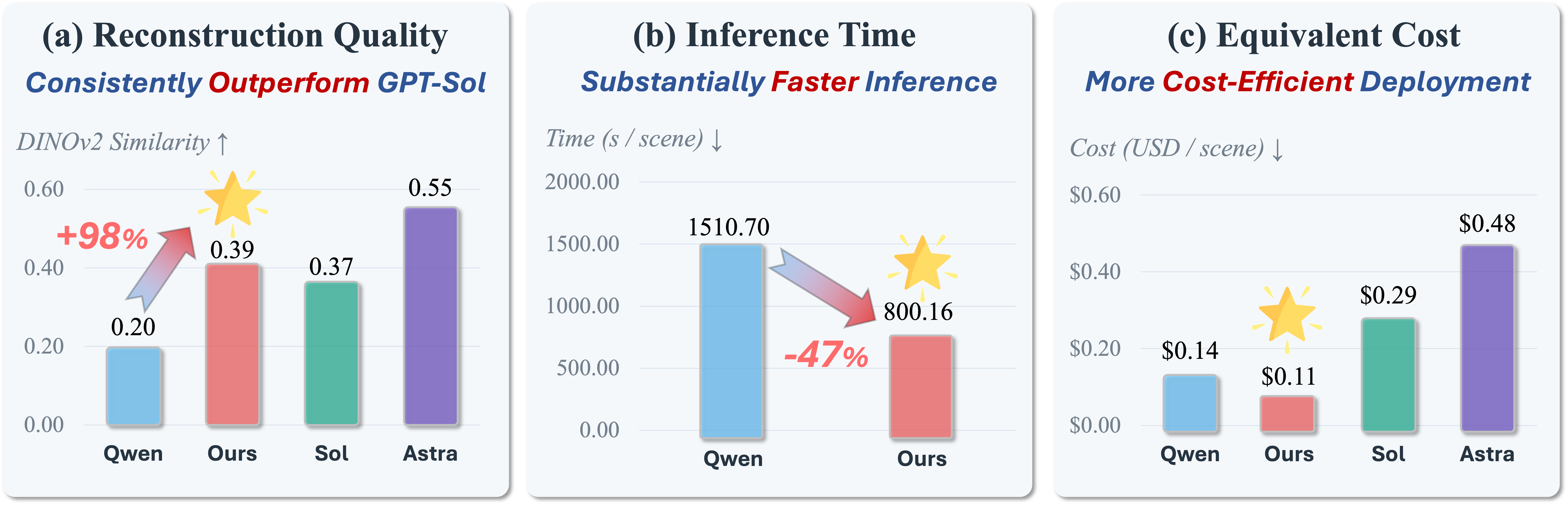}
  \vspace{-0.1cm}
  \caption{Comparison of \method{} with Qwen3.8-27B at \emph{xhigh}, GPT-5.6-Sol at \emph{none}, and GPT-6-Astra at \emph{none}. The average DINOv2-S $\uparrow$, inference time $\downarrow$, and equivalent cost $\downarrow$ are reported.}
  \label{fig:mean-performance}
\end{figure}

\begin{figure}[!t]
  \centering
  \includegraphics[width=0.97\linewidth]{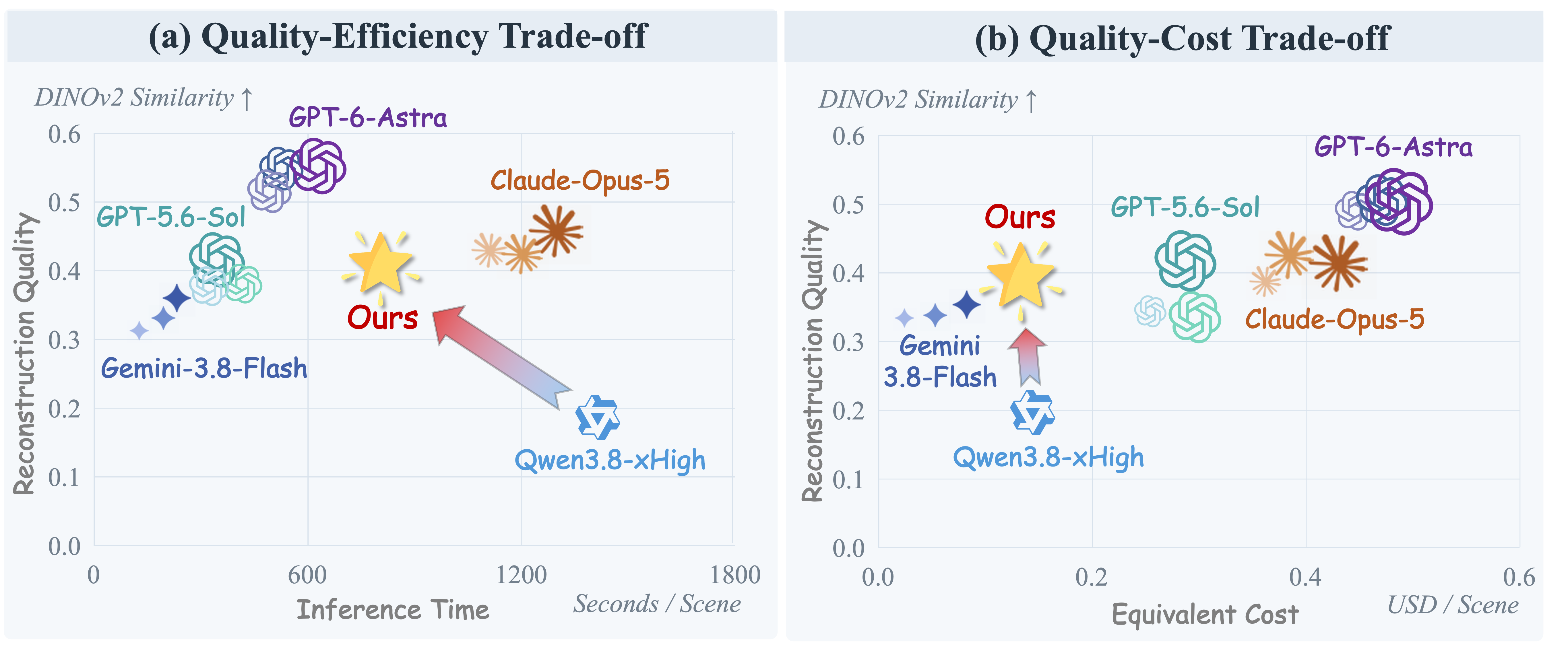}
  \vspace{-0.1cm}
  \caption{Comparison of reconstruction quality against inference time and equivalent deployment cost across model configurations. The \emph{high}, \emph{medium} and \emph{low} settings denote reasoning modes from deeper to shallower inference. DINOv2-S $\uparrow$, inference time $\downarrow$, and equivalent deployment cost $\downarrow$ are reported averaged over three datasets.}
  \label{fig:quality-resource-tradeoff}
  \vspace{0.2cm}
\end{figure}

\begin{figure}[t]
  \centering
  \includegraphics[width=0.97\linewidth]{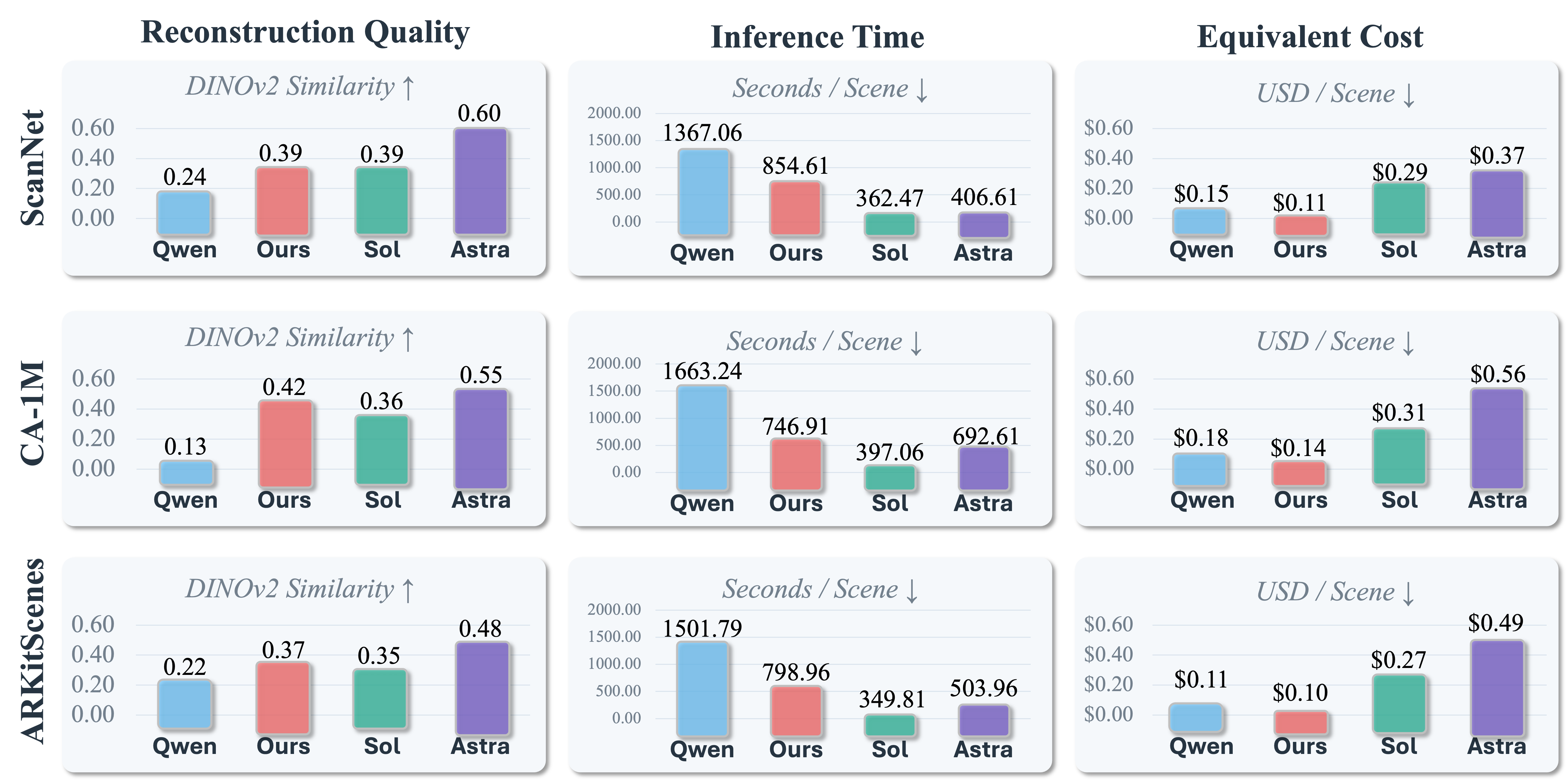}
  \vspace{-0.1cm}
  \caption{Comparison of \method{} with Qwen3.8-27B at \emph{xhigh}, GPT-5.6-Sol at \emph{none}, and GPT-6-Astra at \emph{none} on two in-domain datasets ScanNet and CA-1M and the out-of-domain dataset ARKitScenes. The DINOv2 similarity, inference time, and deployment cost are reported for each.}
  \label{fig:dataset-performance}
  \vspace{0.4cm}
\end{figure}

\subsection{Quantitative Results}

\subsubsection{Comparison with the Base Model}

\paragraph{Reconstruction Quality.}
As shown in Table~\ref{tab:bvb-image-metrics}, \method{} consistently outperforms Qwen3.8-27B across all four reconstruction metrics and three datasets under fixed student weights and reasoning effort. Averaged across datasets, DINOv2-S increases from 0.198 to 0.392, while SSIM improves by 6.7\%, LPIPS decreases by 6.1\%, and PSNR increases by 0.118\,dB; these gains also transfer to the unseen ARKitScenes dataset without target-specific adaptation.

\vspace{-0.2cm}
\paragraph{Inference Time.}
As Figure~\ref{fig:quality-resource-tradeoff} and Figure~\ref{fig:dataset-performance} illustrate, \method{} reduces mean per-scene inference time by 37.5\% on ScanNet, 55.1\% on CA-1M, and 46.8\% on ARKitScenes relative to the base Qwen3.8-27B \emph{xhigh}. Across the three datasets, mean inference time decreases from 1,510.7 to 800.2 seconds, a reduction of 47.0\%, alongside improvements in all four reconstruction metrics.

\vspace{-0.2cm}
\paragraph{Deployment cost.}
As shown in Figure~\ref{fig:quality-resource-tradeoff} and Figure~\ref{fig:dataset-performance}, \method{} achieves lower deployment costs across datasets, reducing the mean equivalent deployment cost by 20.5\% relative to the Qwen3.8-27B xHigh baseline and by 22.7\% on both ScanNet and CA-1M. Overall, \method{} provides a more favorable trade-off between reconstruction quality, inference time, and cost.

\vspace{-0.2cm}
\paragraph{Cross-Model Harness Transfer.} To evaluate the cross-model generalizability of harness evolution, we directly transfer the harness evolved using Qwen3.8-27B to Kimi-K3 under the \emph{high} reasoning-effort setting, without any target-specific search, fine-tuning, or manual adaptation. As shown in Table~\ref{tab:bvb-image-metrics}, the transferred harness consistently improves all four reconstruction metrics across all three datasets compared with the Kimi-K3 baseline without the transferred harness. These results demonstrate that harness evolution can discover reconstruction strategies that generalize beyond the model used during the search, rather than merely exploiting model-specific behaviors.

\subsubsection{Comparison with Frontier Models}

\paragraph{Reconstruction Quality.}
Qwen3.8-27B with \method{} outperforms GPT-5.6-Sol with \emph{none} reasoning effort in eleven of twelve dataset--metric comparisons, improving mean SSIM by 32.3\%, reducing LPIPS by 3.4\%, increasing PSNR by 1.155\,dB and increasing its DINOv2-S by 6.3\%. All four metrics improve on CA-1M and ARKitScenes. Mean SSIM also exceeds Sol-low, Sol-medium, and Sol-high by 32.5\%, 26.3\%, and 20.0\%, respectively. Against Astra-low and Astra-medium, mean SSIM improves by 9.6\% and 7.3\%, respectively.

\paragraph{Inference Time.}
As shown in Figure~\ref{fig:quality-resource-tradeoff} and Figure~\ref{fig:dataset-performance}, Qwen3.8-27B with \method{} averages 800.2 seconds, compared with 369.8 for Sol-none, 502.3 for Astra-low, and 526.8 for Astra-medium. It therefore requires 116.4\%, 59.3\%, and 51.9\% more time, respectively.

\paragraph{Deployment Cost.}
As Figure~\ref{fig:quality-resource-tradeoff} and Figure~\ref{fig:dataset-performance} illustrate, for Qwen3.8-27B with \method{}, the mean equivalent deployment cost decreases by 59.8\% relative to Sol-none, from approximately \$0.290 to \$0.117 per scene. Costs are also 46.2\%, 52.9\%, 69.0\%, and 70.5\% lower than Sol-low, Sol-medium, Astra-low, and Astra-medium, respectively.

\paragraph{Cross-model Harness Transfer.}
To assess the transferability of the evolved harness across model
backbones, we directly apply the harness optimized with Qwen3.8-27B
to Kimi-K3 under the \emph{high} reasoning-effort setting, without
target-specific search, fine-tuning, or manual adaptation. Its reconstruction
performance is competitive with GPT-6-Astra. Meanwhile, the transferred
harness consistently outperforms Claude-Opus-5 across almost all metrics on all three datasets, even across the different reasoning
effort settings evaluated. These results demonstrate that harness evolution can discover
reusable reconstruction strategies that transfer across models, enabling a different model to achieve frontier-competitive
reconstruction quality without model-specific optimization.

\begin{table}[t]
\centering
\caption{Component-ablation on Qwen3.8-27B, averaged across the three datasets. PSNR, SSIM, LPIPS, DINOv2 similarity, inference time, and equivalent cost are averaged across the three datasets.}
\label{tab:ablation}
\begingroup
\small
\setlength{\tabcolsep}{3.5pt}
\renewcommand{\arraystretch}{1.3}
\setlength{\aboverulesep}{0.3ex}
\setlength{\belowrulesep}{0.4ex}
\setlength{\heavyrulewidth}{0.8pt}
\setlength{\lightrulewidth}{0.4pt}

\begin{tabular}{lrrrrrr}
\toprule
Configuration
& PSNR$\uparrow$
& SSIM$\uparrow$
& LPIPS$\downarrow$
& DINOv2-S$\uparrow$
& Time$\downarrow$
& Cost$\downarrow$ \\
\midrule
Base Qwen3.8-27B
& 9.049 & 0.397 & 0.788 & 0.198 & 1510.7 & 0.147 \\
w/o \interface{}
& 9.102 & 0.405 & 0.771 & 0.305 & 1035.0 & 0.152 \\
w/o \harness{}
& 9.119 & 0.411 & 0.762 & 0.285 & 1424.0 & 0.155 \\
\midrule
\rowcolor{violet!8}
\textbf{\method{}}
& \textbf{9.166} & \textbf{0.424} & \textbf{0.739}
& \textbf{0.392} & \textbf{800.2} & \textbf{0.117} \\
\bottomrule
\end{tabular}
\endgroup
\vspace{0.4cm}
\end{table}

\subsection{Qualitative Results}
\label{sec:qualitative}

\begin{figure}[t]
  \centering
  \includegraphics[width=\linewidth]{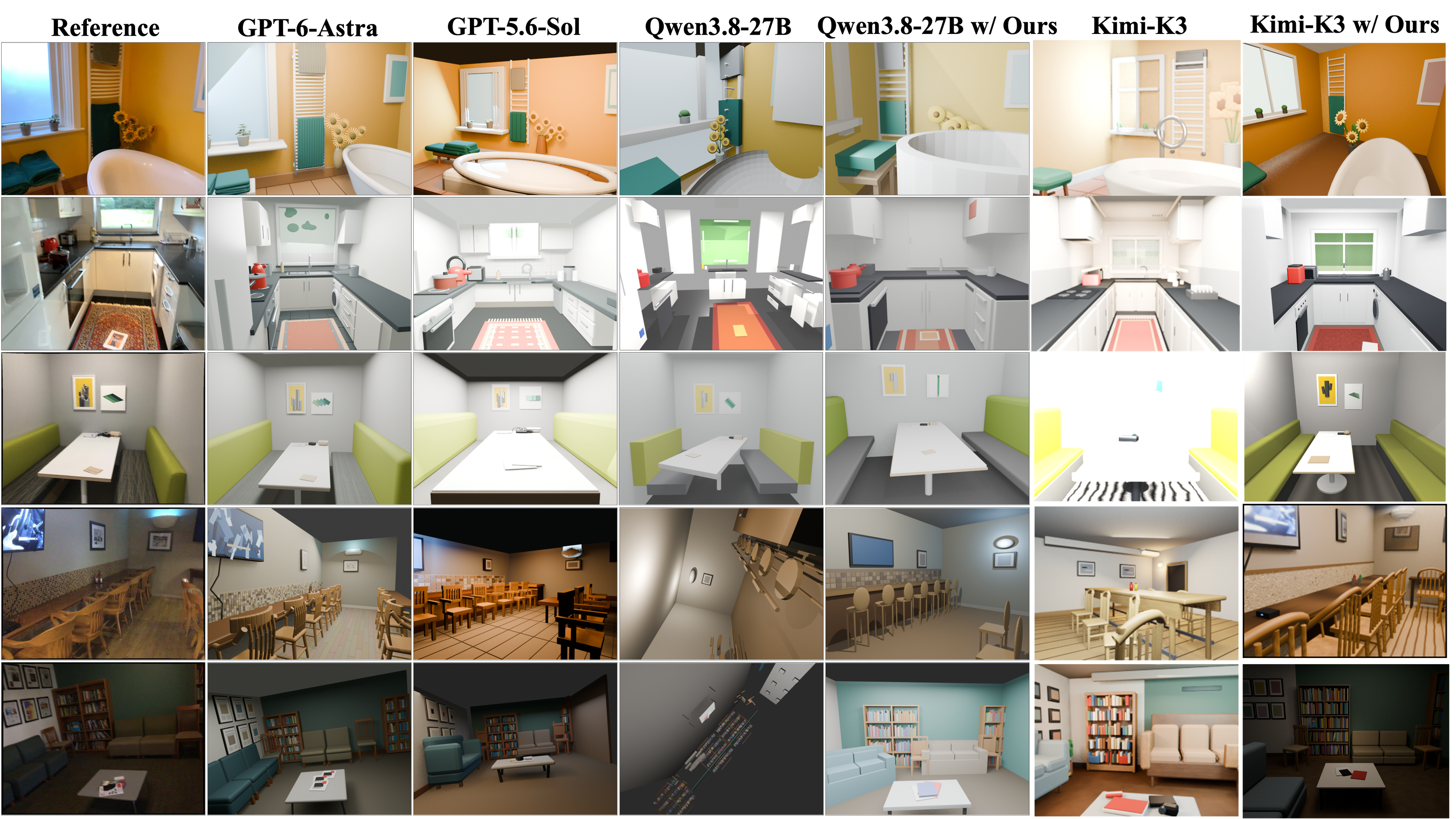}
  \vspace{-0.4cm}
  \caption{Qualitative comparisons of indoor scene reconstruction with GPT-6-Astra, GPT-5.6-Sol, Qwen3.8-27B and Kimi-K3 on Scannet, CA-1M and ARKitScenes datasets. \method{} reconstructs editable 3D scenes that capture object geometry, relative proportions, and spatial relationships in the reference images, demonstrating improved structural fidelity.}
  \label{fig:comparison}
  \vspace{0.5cm}
\end{figure}

\paragraph{Comparison with the Base Model.} As shown in Figure~\ref{fig:comparison}, it compares reconstructions of five indoor scenes in terms of viewpoint alignment, object geometry, and spatial organization. Compared with the base, Qwen3.8-27B with \method{} improves scene coherence, particularly in the bar and lounge, correcting severe viewpoint distortions and better recovering furniture arrangements. In the kitchen, it produces more continuous cabinetry and countertops, avoiding the fragmented cabinet fronts and bright green window region in the baseline. Despite these improvements, discrepancies remain, including an overly bright lounge and simplified seating with inaccurate colors in the booth.

\paragraph{Comparison with Frontier Models.}
Figure~\ref{fig:comparison} also compares both harness-equipped backbones with GPT-6-Astra and GPT-5.6-Sol. For Qwen3.8-27B with \method{}, the bar example shows competitive layout recovery: our reconstruction places a long counter and a repeated row of stools along the wall, whereas GPT-5.6-Sol distributes multiple rows of tables and chairs across the room. In the lounge example, Qwen3.8-27B with \method{} recovers the principal furniture categories and broad arrangement, but renders the sofas and walls in lighter colors than the reference. 

\paragraph{Comparison of Cross-Model Harness Transfer.} The last two columns of Figure~\ref{fig:comparison} compare Kimi-K3 with and without \method{}, using a harness evolved on Qwen3.8-27B. The Kimi-K3 alone exhibits overexposure, bright yellow seating, and a high-contrast floor, whereas the transferred harness better reproduces the reference's colors, furniture, and spatial layout. It also aligns wooden seats with the bar counter and preserves the lounge's dark green accent wall and furniture arrangement. Compared with frontier models, it matches GPT-6-Astra's counter-side seating and booth composition while avoiding the extra furniture rows in GPT-5.6-Sol and better preserving tabletop proportions. Notably, its lounge reconstruction retains the dark green wall more distinctly than either frontier model. These results demonstrate that cross-model harness transfer improves visual fidelity and enables competitive scene-layout and object-detail reconstruction.

\subsection{Ablation Study}
\label{sec:ablation}

We ablate \interface{} and \harness{} with Qwen3.8-27B at \emph{xhigh}. \emph{w/o \interface{}} replaces \interface{} with direct Blender operations under the same optimization budget, while \emph{w/o \harness{}} keeps \interface{} but uses the initial, unevolved harness. The full method combines both.

\paragraph{Effect of \interface{}.}
As shown in Table~\ref{tab:ablation}, incorporating \interface{} increases DINOv2-S from 0.305 to 0.392 and reduces mean inference time by 22.7\% compared with the \emph{w/o \interface{}} variant. These results suggest that the structured interface, which supports hierarchical construction and semantic placement, improves both reconstruction quality and inference efficiency.

\paragraph{Effect of \harness{}.}
As shown in Table~\ref{tab:ablation}, with \interface{} fixed, harness evolution increases DINOv2-S from 0.285 to 0.392, reduces mean inference time by 43.8\%, and lowers the mean equivalent cost per scene from \$0.155 to \$0.117. These results support the value of optimizing a reusable reconstruction procedure alongside the structured action interface.

\section{Conclusion}
\label{sec:conclusion}
We introduced \method{}, a framework for editable, compositional 3D scene reconstruction in code-generated scene settings that combines a structured scene action interface \interface{} with teacher-guided evolution of a reusable reconstruction harness \harness{}. Extensive experiments demonstrate that \method{} improves over its base configuration and outperforms GPT-5.6-Sol with \emph{none} reasoning effort, while reducing mean inference time by 47.0\% and equivalent deployment cost per scene by 59.8\%. Furthermore, a reconstruction harness evolved using Qwen3.8-27B can also generalize to different models like Kimi-K3, achieving reconstruction results competitive with those of GPT-6-Astra.
\label{maintextend}

\bibliography{references}
\bibliographystyle{iclr2027_conference}

\clearpage
\appendix
\section{Experiment Details}

\subsection{Dataset Details}
To evaluate the effectiveness of \method{}, we conduct experiments on three representative indoor-scene datasets covering different capture settings: ScanNet~\citep{dai2017scannet}, CA-1M~\citep{lazarow2024cubify}, and ARKitScenes~\citep{baruch2021arkitscenes}. ScanNet and CA-1M serve as in-domain datasets for harness development and evaluation, while ARKitScenes is reserved for out-of-domain evaluation. We use 160 scenes in total, comprising 100 development scenes, 40 in-domain test scenes, and 20 out-of-domain test scenes. All methods are evaluated on the same held-out scenes under a consistent single-image reconstruction protocol. Input views are selected using a space-aware frame sampling strategy following Spatial-MLLM~\citep{wu2025spatialmllm}. For each in-domain dataset, all views and repeated captures of the same physical scene are kept within a single split to prevent development--test overlap. Detailed descriptions are provided below.

\begin{itemize}
    \item \textbf{ScanNet.} ScanNet~\citep{dai2017scannet} provides RGB-D sequences of indoor environments with camera poses, reconstructed 3D geometry, and semantic annotations. We randomly sample 70 scenes and divide them into 50 development scenes and 20 held-out in-domain test scenes. Its indoor layouts and object arrangements make it suitable for assessing whether a reconstruction agent can recover coherent scene geometry, constituent objects, and spatial relationships from a single image. The development split supplies reconstruction feedback for harness evolution, whereas the test split is used exclusively for evaluation.

    \item \textbf{CA-1M.} CA-1M~\citep{lazarow2024cubify} contains handheld egocentric indoor captures registered to high-quality laser scans, together with dense 3D object annotations. We randomly sample 70 scenes, assigning 50 to harness development and 20 to held-out in-domain evaluation. Compared with ScanNet, its capture and annotation setting provides a complementary test of object-level geometry, relative proportions, and spatial composition. As with ScanNet, only the development scenes are used to evaluate candidate harness revisions; the held-out test scenes are not used during harness search.

    \item \textbf{ARKitScenes.} ARKitScenes~\citep{baruch2021arkitscenes} consists of mobile RGB-D captures of indoor environments with laser-scanned geometry and 3D object annotations. We randomly sample 20 scenes and use them exclusively for out-of-domain evaluation. Unlike the two in-domain datasets, ARKitScenes contributes no images or reconstruction feedback to harness development. The student parameters, selected harness, and inference configuration are transferred without modification, making this benchmark a test of whether the reconstruction procedure generalizes to previously unseen data.
\end{itemize}

\subsection{Counterpart Details}
This section provides a detailed overview of the counterpart models employed in our experiments. We compare \method{} with the unmodified student backbone and representative proprietary multimodal models, including GPT-5.6-Sol, GPT-6-Astra, Claude-Opus-5, and Gemini-Flash-3.8. The proprietary models are evaluated as zero-shot agents for compositional image-to-3D scene reconstruction without task-specific fine-tuning. For models supporting configurable reasoning effort, each model--effort combination is treated as a separate counterpart. All counterparts are evaluated on the same held-out scenes using consistent reconstruction metrics. Details are provided below.

\begin{itemize}
    \item \textbf{Qwen3.8-27B.} Qwen3.8-27B~\citep{qwen38_27b} serves as the open-source backbone and the primary base-model counterpart. We evaluate it at \emph{xhigh} reasoning effort, matching the model and inference effort used by the frozen student in \method{}. This comparison isolates the improvements obtained from the proposed reconstruction framework without changing the underlying student model parameters.

    \item \textbf{Kimi-K3.}
    Kimi-K3~\citep{kimik3} is an open-weight multimodal model evaluated at \emph{high} reasoning effort to investigate the cross-model generalization of the evolved harness. Specifically, we directly transfer the harness evolved with Qwen3.8-27B to Kimi-K3 without additional harness evolution, model fine-tuning, or target-model-specific optimization. We compare Kimi-K3 equipped with the transferred harness against its unevolved counterpart under the same reasoning effort, reconstruction protocol, and execution budget. The \interface{} interface and execution backend remain unchanged throughout the evaluation. This setting isolates the transferability of the learned reconstruction procedure across different model backbones, examining whether harness improvements obtained from one student model can generalize to another without model-specific adaptation.

    \item \textbf{GPT-5.6-Sol.} GPT-5.6-Sol~\citep{gpt56sol} is a proprietary multimodal model used as a frontier-model counterpart for evaluating reconstruction quality, inference efficiency, and deployment cost. We consider four reasoning-effort configurations: \emph{none}, \emph{low}, \emph{medium}, and \emph{high}. The \emph{none} configuration serves as the primary proprietary reference in our main comparison, while the remaining configurations characterize the effect of increased inference effort on reconstruction performance and resource consumption.

    \item \textbf{GPT-6-Astra.} GPT-6-Astra~\citep{gpt6astra} provides an additional frontier-model reference for compositional scene reconstruction. We evaluate its \emph{low}, \emph{medium}, and \emph{high} reasoning-effort configurations. Separately, GPT-6-Astra at \emph{medium} serves as the teacher model during \harness{} development, proposing harness revisions from development-set feedback. The teacher does not participate in reconstruction at test time.

    \item \textbf{Claude-Opus-5.} Claude-Opus-5~\citep{claudeopus5} is a proprietary multimodal counterpart evaluated under four reasoning-effort settings: \emph{none}, \emph{low}, \emph{medium}, and \emph{high}. It provides an additional comparison of image-conditioned scene-code generation across model families. These configurations are evaluated without task-specific fine-tuning, enabling an assessment of reconstruction fidelity and inference resource requirements.

    \item \textbf{Gemini-Flash-3.8.} Gemini-Flash-3.8~\citep{geminiflash38} is another proprietary multimodal counterpart for single-image scene reconstruction. We evaluate its \emph{low}, \emph{medium}, and \emph{high} reasoning-effort configurations. Together with the other proprietary models, it broadens the comparison across model families and inference configurations, allowing us to examine the trade-offs among quality, inference time, and cost.
\end{itemize}

\section{Additional Related Work}

\subsection{Self-improving agents and harness optimization} Self-improving agent systems leverage feedback from prior executions to iteratively refine the mechanisms that govern subsequent behavior. Existing approaches differ primarily in the components they optimize and the forms of feedback they exploit. Reflexion~\citep{shinn2023reflexion} records natural-language reflections in episodic memory to inform future decisions, whereas STOP~\citep{stop} recursively improves an executable scaffold. A complementary line of work, including TextGrad~\citep{textgrad}, ProTeGi~\citep{protegi}, GEPA~\citep{gepa}, and Feedback Descent~\citep{feedback}, optimizes textual components using natural-language feedback. More recently, Meta-Harness~\citep{metaharness} and Self-Harness~\citep{selfharness} extend the optimization target to agent source code, using execution outcomes to guide iterative code revisions. Their evaluations span diverse tasks: Meta-Harness considers classification, mathematical reasoning, and terminal-based coding, while Self-Harness evaluates performance on Terminal-Bench, SWE-bench, and AppWorld. Although these methods differ in their optimization targets and feedback mechanisms, they share a common objective: improving agent performance through iterative system-level refinement rather than relying exclusively on conventional model training. \method extends this line of research to code-based 3D scene reconstruction, where agent performance depends not only on the underlying model but also on the execution logic that translates reconstruction objectives into executable code and rendered scenes. 

\section{Implementation Details}
\label{app:implementation}
All experiments were conducted on a computing server equipped with four NVIDIA L40S GPUs (48 GB each), an AMD EPYC 9654 CPU (112 cores and 224 threads), and 755 GiB of RAM. The software environment consisted of Ubuntu 24.04.3 LTS, Python 3.10.12, PyTorch 2.5.1+cu124.

\section{Harness Design and Evolution}
\subsection{Initial harness}
We use a minimal initial harness as the shared baseline for all models in our benchmark. For baseline evaluation, every model is assessed under the same harness and prompt protocol to ensure comparability. Each example is processed through a direct model call only; we do not use external coding-agent harnesses or frameworks such as Codex, Claude Code, or Pi. The harness consists exclusively of the system prompt and the user prompt shown below. It does not include MCP tools, skills, an agent-loop architecture, memory, or other external tool-use or orchestration components. For each example, the model receives a single textual prompt together with the reference image and is expected to return a complete, executable Blender Python program. As specified in the system prompt, file saving and rendering are handled by a separate local evaluation script rather than by an agent framework. The exact prompts used in this baseline are provided below.
\newtcolorbox{promptbox}[1][]{
  enhanced,
  breakable,
  colback=gray!12,
  colframe=gray!55,
  boxrule=0.5pt,
  arc=2pt,
  left=8pt,
  right=8pt,
  top=6pt,
  bottom=6pt,
  title=#1,
  coltitle=black,
  fonttitle=\bfseries,
}

\begin{promptbox}[System Prompt]
You are an expert in reconstructing indoor scenes from reference images using Blender Python. Return only complete, executable Python code, with no explanations or Markdown. Accurately recreate the scene using actual 3D geometry, materials, lighting, and a camera. Do not use the input image as a background or substitute for 3D modeling. Explicitly set the active scene camera. Handle scene modeling and configuration only. Do not save files or render the scene, as these operations will be handled by a separate local script.
\end{promptbox}

\begin{promptbox}[User Prompt]
Reconstruct the indoor scene from the input image using Blender 3D modeling. Preserve the original spatial layout, architectural structures, furniture, objects, proportions, positions, orientations, and occlusion relationships. Strictly match the original camera viewpoint, perspective, composition, and framing. Retain the original materials, colors, and textures. Do not redesign the scene or invent off-frame content. The input image is <image>.
\end{promptbox}

\subsection{The evolved final harness}

\textbf{Planning Stage.} Given the reference image and reconstruction instruction, the harness first analyzes the scene to identify visible architectural structures, furniture, and object instances, estimating their geometry, proportions, appearance, camera viewpoint, and key spatial relationships, including relative positioning and occlusion. Based on this analysis, it constructs a hierarchical scene plan that organizes the scene into individual assets and their constituent parts, while specifying relevant object attributes, spatial configurations, and inter-object relationships. The resulting plan is then serialized into a structured JSON document that serves as a machine-readable intermediate representation for downstream code generation. By explicitly encoding the scene hierarchy, object properties, and spatial constraints, this document provides a clear and consistent specification for the subsequent code generation stage. Throughout the planning process, the harness grounds its decisions in visible evidence, treating uncertain or occluded regions conservatively and avoiding unsupported assumptions.

\textbf{Code Executing Stage.} The coding agent translates the scene plan into executable Blender Python through \interface{} interface. Parameterized primitives represent local geometry, the asset--part hierarchy captures object structure, and semantic spatial operators express relations such as alignment, distribution, support, and attachment. The harness encourages the agent to construct objects compositionally, reuse consistent reference frames, and specify the camera, materials, colors, and lighting explicitly. This organization reduces reliance on independently guessed low-level coordinates and makes it possible to localize later edits to particular objects or parts.

\textbf{Inspect and Render Feedback.} The harness evaluates the generated scene through iterative execution, rendering, and inspection. After executing the Blender script, the harness first checks for syntax errors, runtime failures, and other execution issues. If the script runs successfully, the resulting scene is rendered from the target camera viewpoint and inspected against the reference image and the structured scene plan. The inspection combines visual assessment with spatial validation: a vision-language model evaluates global composition, camera framing, object appearance, materials, colors, lighting, and occlusion patterns, while spatial checks examine object dimensions, relative positions, alignment, support, attachment, and unintended intersections where applicable. Based on the identified issues, the harness generates targeted repair instructions associated with specific objects, parts, or scene attributes, allowing the coding agent to modify the relevant portions of the Blender code rather than regenerate the entire scene. The updated code is then executed and rendered again, forming an iterative feedback loop in which each revision is evaluated against the reference and the original scene plan. By grounding code revisions in both rendered evidence and explicit spatial constraints, this stage improves visual fidelity and structural consistency while reducing unnecessary modifications to components that already satisfy the reconstruction requirements.

\section{Limitation}
\method provides an effective approach to harness optimization, while leaving room for further improvement. First, harness evolution incurs substantial computational costs due to repeated teacher-guided analysis, candidate generation, scene reconstruction, rendering, and evaluation, highlighting a trade-off between harness quality and search efficiency. Second, harness optimization improves task execution but does not directly enhance the underlying model's capabilities, and the evolved system does not consistently outperform strong models such as GPT-6-Astra. Future work could jointly evolve RSI harnesses and adapt RSI models to improve reconstruction quality.

\end{document}